\documentclass{article}

\usepackage{microtype}
\usepackage{graphicx}
\usepackage{subfigure}
\usepackage{booktabs}

\usepackage{hyperref}

\usepackage[accepted]{style}

\usepackage{amsmath}
\usepackage{amssymb}
\usepackage{mathtools}
\usepackage{amsthm}

\usepackage[capitalize,noabbrev]{cleveref}

\theoremstyle{plain}

\theoremstyle{definition}

\theoremstyle{remark}

\usepackage[textsize=tiny]{todonotes}
\conftitlerunning{Gradient-based Design of Computational Granular Crystals}

\begin{document}

\twocolumn[
\conftitle{Gradient-based Design of Computational Granular Crystals}

\begin{confauthorlist}
\confauthor{Atoosa Parsa}{uvm}
\confauthor{Corey S.~O'Hern}{yale}
\confauthor{Rebecca Kramer-Bottiglio}{yale}
\confauthor{Josh Bongard}{uvm}
\end{confauthorlist}

\confaffiliation{uvm}{Department of Computer Science, University of Vermont, Burlington, VT, USA}
\confaffiliation{yale}{Department of Mechanical Engineering and Materials Science, Yale University, New Haven, CT, USA}
\confcorrespondingauthor{Atoosa Parsa}{atoosa.parsa@uvm.edu}

\confkeywords{Deep Physical Neural Networks, Unconventional Computing, Granular Crystals}

\vskip 0.3in
]

\printAffiliationsAndNotice{}

\begin{abstract}
There is growing interest in engineering unconventional computing devices that leverage the intrinsic dynamics of physical substrates to perform fast and energy-efficient computations. Granular metamaterials are one such substrate that has emerged as a promising platform for building wave-based information processing devices with the potential to integrate sensing, actuation, and computation. Their high-dimensional and nonlinear dynamics result in nontrivial and sometimes counter-intuitive wave responses that can be shaped by the material properties, geometry, and configuration of individual grains. Such highly tunable rich dynamics can be utilized for mechanical computing in special-purpose applications. However, there are currently no general frameworks for the inverse design of large-scale granular materials. Here, we build upon the similarity between the spatiotemporal dynamics of wave propagation in material and the computational dynamics of Recurrent Neural Networks to develop a gradient-based optimization framework for harmonically driven granular crystals. We showcase how our framework can be utilized to design basic logic gates where mechanical vibrations carry the information at predetermined frequencies. We compare our design methodology with classic gradient-free methods and find that our approach discovers higher-performing configurations with less computational effort. 
Our findings show that a gradient-based optimization method can greatly expand the design space of metamaterials and provide the opportunity to systematically traverse the parameter space to find materials with the desired functionalities.
\end{abstract}

\section{Introduction}
\label{sec:introduction}
The widespread adoption of Artificial Neural Networks in AI applications can be traced back to the early '90s when the seminal work of Hornik et al. proved them to be universal function approximators \yrcite{Hornik.etal1989}. The efficient execution of the backpropagation algorithm on Graphics Processing Units enabled the training of large Deep Neural Networks (DNNs) and initiated a new era where the focus was on the development of data-intensive methodologies that leverage available computing power \cite{Krizhevsky.etal2012}. However, the unprecedented growth of power requirements and the recent slowdown of Moore's law have made it challenging for the semiconductor industry to keep up with computing demand \cite{Lemme.etal2022a}. This has motivated the development of alternate information processing platforms that relax the classic assumptions on the computing models, architectures, and substrate choices in favor of fast and efficient execution of special-purpose computations. 

Advances in physics, chemistry, and materials science, along with revolutionary fabrication and manufacturing technologies, have provided the opportunity to explore unconventional computing paradigms that abandon the notion of centralized processing units and harness the natural dynamics of the physical system to perform the desired computation. With this perspective, any controllable physical system with rich intrinsic dynamics can be exploited as a computational resource. This has resulted in the development of mechanical \cite{Lee.etal2022a}, optical \cite{Anderson.etal2023}, electromechanical \cite{ElHelou.etal2022a}  and biological \cite{Roberts.Adamatzky2022} computing units. Such physics-based computing devices offer potential advantages for fast and efficient computation that avoids analog-to-digital conversion and allows massively parallel operations \cite{Yasuda.etal2021}. However, finding the best hardware setup is often a challenging task beyond the intuitive limits of human experts and can benefit from automatic design methodologies to tune various aspects of the system according to the application \cite{Finocchio.etal2023}. 

This paper focuses on the inverse design methodologies for computational metamaterials. Metamaterials are engineered composite materials designed with particular spatial configurations that exhibit macroscopic behaviors different from their constituent parts \cite{Xia.etal2022}. They can possess non-natural static or dynamic properties such as negative bulk moduli and mass density, non-reciprocity, and auxetic behavior \cite{Kadic.etal2019, Jiao.etal2023}. Mechanical metamaterials, especially those made of field-responsive materials, have received immense attention for robotics applications where they can respond to various stimuli and reconfigure to adapt to different environmental conditions \cite{Rafsanjani.etal2019}. They provide increased robustness and reduced power consumption in the system and enable the design of highly tunable multifunctional mechanisms that integrate sensing, information processing, and actuation in fully autonomous engineered systems \cite{Pishvar.Harne2020}. The ability to create metamaterials that can manipulate mechanical vibrations of varying frequencies has made them an excellent platform for mechanical computation \cite{Yasuda.etal2021}.

Here, we concentrate on a subset of metamaterials with particulate structures, namely granular crystals. These are composite materials made of noncohesive particles with various material properties and shapes, which are densely packed in random or carefully designed configurations \cite{Karuriya.Barthelat2023}. Due to their discrete nature and the nonlinearity of interparticle contacts, granular materials exhibit highly tunable dynamic responses, which are of great interest to both the academic community and industrial organizations. They are utilized in a broad range of applications, including energy localization and vibration absorption layers \cite{Zhang.etal2015, Taghizadeh.etal2021b}, acoustic computational units like switches and logic elements \cite{Li.etal2014a, Parsa.etal2023}, granular actuators \cite{Eristoff.etal2022}, acoustic filters \cite{Boechler.etal2011}, and sound focusing/scrambling devices \cite{Porter.etal2015a}. Beyond practical applications, granular assemblies are also studied as simple test beds for investigating fundamental phenomena in many disciplines, including materials science and condensed-matter physics \cite{Rodney.etal2011, Jaeger.Nagel1992}.

Granular crystals are commonly studied in a confined structure subject to external vibrations. Their nonlinear dynamic response is highly tunable by local changes to individual particles' properties. Therefore, they possess great potential for wave-based physical computation. However, with such high-dimensional parameter space and strongly nonlinear discrete dynamics, tuning their vibrational response is extremely challenging. Many studies are limited to experimental measurements \cite{Boechler.etal2011, Li.etal2014a, Lawney.Luding2014, Cui.etal2018} or numerical integration of the equations of motion \cite{Boechler.etal2011, Chong.etal2017}. Analytical methods for such granular systems primarily focus on reduced-order linearized approximations. 
In most such investigations, the discrete nature of the system is ignored, and the system is analyzed in the continuum limit.  
However, such analysis fails to capture nonlinear phenomena that emerge from the non-integrable discreteness in the system and the response can diverge significantly from the predictions \cite{Somfai.etal2005, Nesterenko.etal2005}. 
Therefore, there is currently no general systematic methodology for studying the temporal and spatial characteristics of the wave response in disordered granular crystals and designing materials with the desired dynamic response \cite{Ganesh.Gonella2017}.

In this paper, we develop a differentiable simulator for granular crystals that can be incorporated into an optimization pipeline to find the best material properties to perform mechanical computations.

\section{Related Work}
\label{sec:relatedwork}
Physical computing has been an active research topic in recent years, and many attempts have been made to take inspiration from deep learning concepts and incorporate machine learning techniques in designing novel computational hardware. Physical Reservoir Computing (PRC) is one such direction where a physical system is exploited for computation by applying the inputs to a physical substrate, collecting the raw measurements, and only training a linear ``readout'' layer to match the desired outputs \cite{Nakajima2020}. Recently, Physical Neural Networks (PNNs) have been introduced in which the hardware's physical transformation is trained in a similar manner to DNNs to perform the desired computations. Here, unlike PRC, the system's input-output transformation is directly trained with an algorithm called physics-aware training (PAT) that enables backpropagation on physical input and output sequences \cite{Wright.etal2022}. Optical Neural Networks are an example of PNNs, that propose running deep learning frameworks for any task, such as classification or natural language processing, directly on an optical hardware instead of a digital electronic one such as a GPU \cite{Anderson.etal2023, Huo.etal2023}. In PNNs, the mechanical properties of the physical system do not change during the training; instead, the applied physical input is tuned with backpropagation using a differentiable model of the system to achieve the desired input-output transformation. Training PNNs with BP has a couple of drawbacks such as needing accurate knowledge of the physical system and being unsuitable for online training. Direct feedback alignment (DFA) was developed to address this by omitting the need for layer-by-layer propagation of error. However, it still requires modeling and simulation of the physical system \cite{Nakajima.etal2022}.

Mechanical Neural Networks (MNNs) are another type of physical network that, unlike the previous works, tune the mechanical properties of the physical system during training. Lee et al. have developed a framework where the stiffness values of interconnected beams in a lattice are tuned for desired bulk properties like shear and Young's modulus or mechanical behaviors such as shape morphing \yrcite{Lee.etal2022a}. Similar works have been done for analog wave-based computing where a differentiable model is developed based on the finite difference discretization of the dynamical equations describing a scalar wave field in continuous elastic metamaterials \cite{Hughes.etal2019, Jiang.etal2023a}. The same approach is utilized in \cite{Papp.etal2021} for designing computing devices with spin waves propagating in a magnetic thin film. Here, a magnetic field distribution is designed to steer the spin waves in order to achieve the desired behavior. However, the system is not trained in hardware; instead, the material is discretized into cells with various material properties, that are determined using an approximate differentiable simulator. Such an approximate model will not capture the full dynamical behavior in the strongly nonlinear regime. Moreover, after manufacturing the optimized design there are no methods for online adaptation of the structural parameters and therefore such physical substrates are more suitable for tasks where a system is trained once and then used for inference many times.

Granular metamaterials are particulate systems where the properties of the individual particles can be modified independently. Therefore they offer the opportunity to build reconfigurable multifunctional materials. While gradient-based optimization has been explored to a great extent in designing continuous photonic materials \cite{Tahersima.etal2019, Yao.etal2019, Mao.etal2021, Jiang.etal2021}, designing granular crystals with desired dynamic responses has not been explored. There exists an extensive body of research on granular materials dating back over $200$ years. However, a general connection between their dynamic wave response and their constituents' shapes and material properties remains unknown. Moreover, analytical exploration of the parameter space of granular materials is infeasible without imposing simplifying assumptions and approximations. In this paper, we present a gradient-based optimization framework for designing granular crystals with desired dynamic wave responses.
In summary, we make the following contributions:
\begin{itemize}
    \item Present a gradient-based optimization platform for designing granular crystals with a desired dynamic response without recourse to any continuum approximations of the physics model.
    \item Demonstrate the application of our proposed framework for designing wave-based mechanical computing devices.
    \item Compare the performance and computational efficiency of our proposed method to gradient-free optimization methods incorporated in previous related works.
\end{itemize}

\section{Methods}
\label{sec:methods}
\begin{figure*}[ht]
\vskip -0.1in
\begin{center}
\centerline{\includegraphics[width=\textwidth]{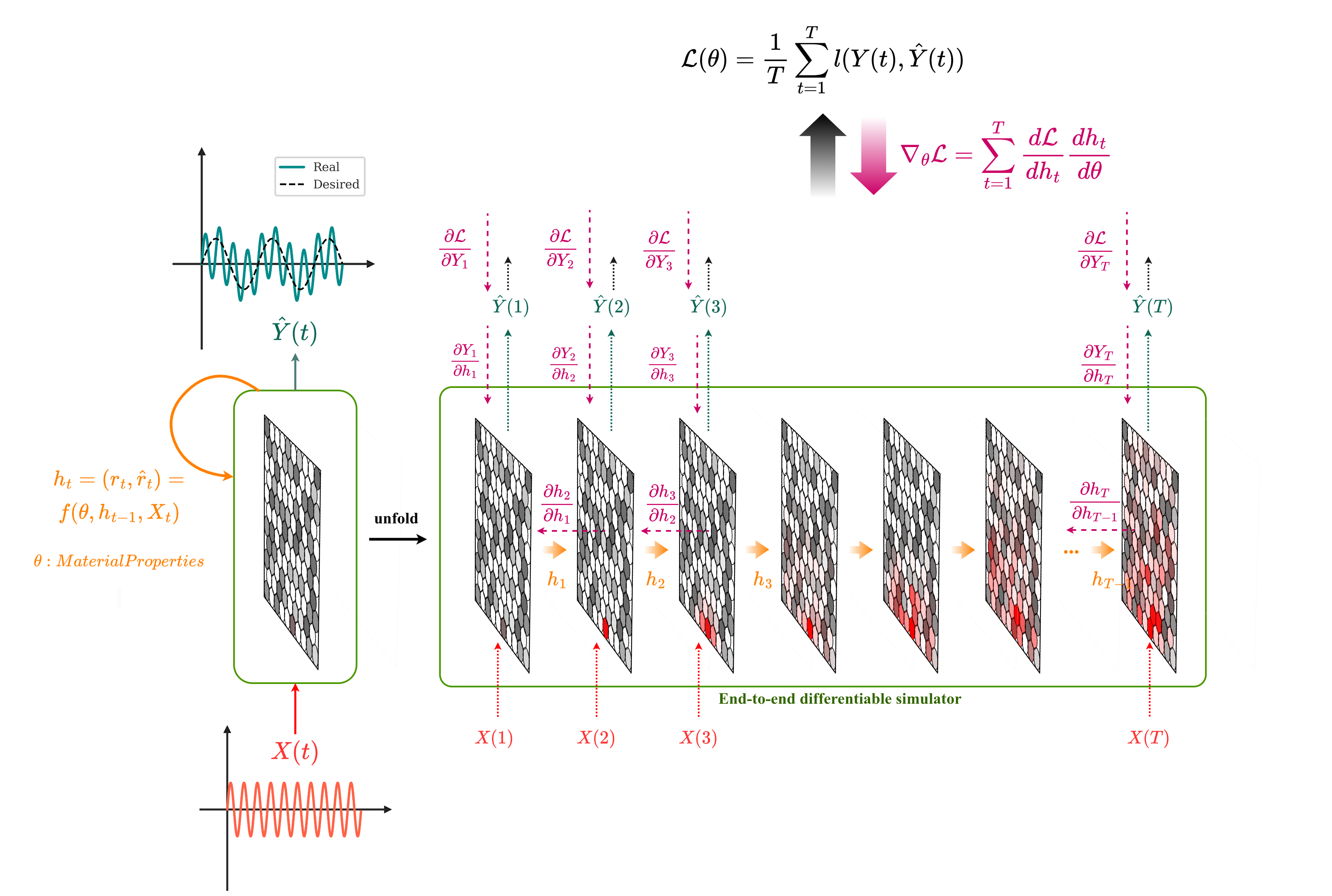}}
\caption{Inverse design of computational granular crystals. When the granular crystal is vibrated at its boundary, the elastic compression waves (indicated by the red shades in the panels) propagate in the material until they are scattered or attenuated by disorder, affected by dispersion, or distorted by (self-)demodulation and frequency mixing at nonlinear interparticle contacts (\textit{Forward Pass}). The waves arriving at the output particle(s) are recorded and the difference between the desired ($Y(t)$) and recorded response ($\hat{Y}(t)$) is utilized in a loss function ($\mathcal{L}$) to adjust the trainable parameters ($\theta$). $f$ relates the input $X_t$, parameters $\theta$, and the hidden state of the system $h_{t-1}$ to the hidden state at the next time step $h_t$. An end-to-end differentiable physics simulator allows us to track the partial derivatives in the \textit{Backward Pass} indicated by the pink arrows in the figure. The particles' material properties can be optimized with a gradient-based method to produce the desired nonlinear wave response.}
\label{fig:overview}
\end{center}
\vskip -0.1in
\end{figure*}

\cref{fig:overview} shows an overview of our optimization framework. A dense packing of circular particles with different material properties is subjected to external mechanical vibrations by displacing the selected input particle(s) with a predefined oscillatory force indicated as $X(t)$. The system's hidden state ($h_t={(r, \hat{r})}_t$) can be described with the position ($r_t$) and velocity ($\dot{r}_t$) of the particles in time. In the forward pass, the system's state evolves according to the dynamics dictated by the physical system and depends on the state in the previous time step ($h_{t-1}$), physical parameters ($\theta$), and the input at time $t$ ($X_t$). The physics model describes the nonlinear relation between the state, input, and the parameters as $h_t = f(\theta, h_{t-1}, X_t)$. This is analogous to Recurrent Neural Networks (RNNs), where the hidden state allows the network to remember the past information fed into the network and enables learning of the temporal structure and long dependencies in the input.

The output is defined as the measurements of a physical property of the system in time such as the displacement of the chosen output particle(s) $\hat{Y}_t$. To train the physical network, we need to update the trainable parameters $\theta$, which are the material properties of the particles (equivalent to weights of an RNN), to reduce the loss $\mathcal{L}$ defined between the real $\hat{Y}_t$ and desired $Y(t)$ outputs over $T$ time steps.   

An end-to-end differentiable simulator allows us to retain the gradients of the loss function with respect to the trainable parameters ($\nabla_{\theta} \mathcal{L}$) to be used in backpropagation. Similar to traditional RNNs, the gradients are obtained by taking the partial derivatives and using the chain-rule as follows:
\begin{align}
    & \nabla_{\theta} \mathcal{L} = \sum_{t=1}^{T} \frac{d \mathcal{L}}{d h_t} \frac{d h_t}{d \theta} \nonumber \\
    & \frac{d \mathcal{L}}{d h_t} = \frac{\partial \mathcal{L}}{\partial h_t} + \frac{\partial \mathcal{L}}{\partial h_{t+1}} \frac{\partial h_{t+1}}{\partial h_t} \nonumber \\
    & \frac{\partial \mathcal{L}}{\partial h_t} = \frac{\partial \mathcal{L}}{\partial Y_t} \frac{\partial Y_t}{\partial h_t} \nonumber \\
    & \frac{d h_t}{d \theta} = \frac{\partial f}{\partial \theta} \label{eq:grads_BP}
\end{align}
Having the gradients of the loss function with respect to the physical parameters of the network, a gradient-based optimization method can be used to update the parameters ($\theta$) at time step $T$ and start the next forward pass. In the next section, we first outline the differentiable simulator that was developed for granular crystals. We then provide the specifics of our optimization pipeline in \cref{sec:GD}.

\subsection{Differentiable Simulator}
\label{sec:simulator}
\begin{figure}[ht]
\vskip 0.1in
\begin{center}
\centerline{\includegraphics[width=1.0\columnwidth]{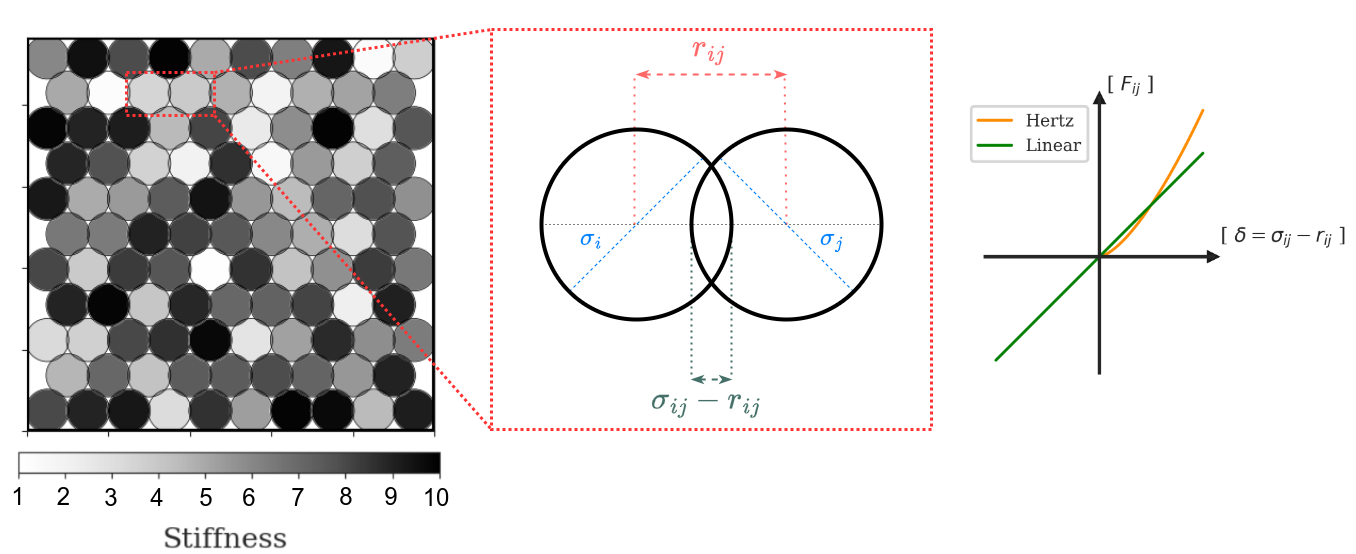}}
\caption{A granular crystal is made of spherical particles with identical size (diameter $\sigma$) and various stiffnesses (represented in different shades of grey) in a confined configuration with fixed boundaries. \textit{Hertz's law} describes the relation between the particle's overlap ($\delta=\sigma_{ij}-r_{ij}$) and applied force ($F$) as $F=\alpha \delta ^ \beta$. Here, $\beta$ is a constant that depends on the particle geometry and determines the nonlinearity of the contact forces. A commonly used value for spherical contacts is $\beta = \frac{3}{2}$, which produces a cubic nonlinearity in the equations of motion. $r_{ij}=|r_i-r_j|$ is the interparticle distance and $\sigma_{ij}=\frac{\sigma_i+\sigma_j}{2}$ is the maximum distance, after which the particles lose contact. As it is shown in the plot on the right, the interparticle potential is one-sided, and unlike a Hookean spring, the force becomes zero when the two particles lose contact.}
\label{fig:model}
\end{center}
\vskip -0.1in
\end{figure}

\cref{fig:model} shows an overview of the granular crystals we aim to optimize in this work. Deformable spherical particles with identical diameters and various elasticity are placed on a hexagonal lattice with fixed boundaries in both $x$ and $y$ directions. In this system, the repulsive force between two neighboring particles is nonlinear and can be described by the Hertz law \cite{Hertz1882}. More details about the physics model are provided in Appendix~\ref{sec:model}.

the Discrete Element Method (DEM) \cite{Cundall.Strack1979} can be used to numerically simulate the motion of the interacting particles in a granular crystal. In this paper, we developed a differentiable simulator with the same method (see \cref{sec:simulation} for more details) in the \textit{PyTorch} framework \cite{Paszke.etal2017}.

\subsection{Optimization Setup}
When a disordered granular crystal, such as the one shown in \cref{fig:model}, is vibrated at its boundaries, the produced elastic waves propagate through the material and scatter at the particle-particle interfaces. The material properties of the individual particles (elasticity, density, etc.), their geometry (shapes and sizes), and their arrangement (neighboring contact points) determine the distortion of the elastic waves and their frequency and amplitude-dependent attenuation. In this paper, we formulate the optimization problem as finding the stiffness values of the particles in a hexagonal granular crystal to achieve a desired wave response. Therefore, the trainable parameters, as defined in \cref{eq:grads_BP}, are $\theta={k_i, i \in [0, N]}$ where $N$ is the total number of the particles. The desired wave response is defined in terms of the displacement of the selected output particles and formulated into the loss function $\mathcal{L}$. The details of the desired output and the loss function will be provided in \cref{sec:results}. 

\subsection{Gradient-based Optimization}
\label{sec:GD}
To enable the gradient-based optimization of granular crystals, we used PyTorch's automatic differentiation (\textit{autodiff}) engine to compute the gradients of the loss function with respect to the trainable material properties ($\theta$). We implemented custom submodules for the granular crystal simulation. Adam optimizer \cite{Kingma.Ba2017} with an adaptive learning rate is utilized for the training process. The loss function and training parameters for each experiment are indicated in \cref{sec:results}

\subsection{Gradient-free Optimization}
\label{sec:EA}
Evolutionary algorithms are a class of population-based gradient-free optimization methods that are well-suited for searching the high-dimensional parameter spaces and tackling multi-objective black-box design problems. They are particularly powerful methods for problems with extremely rugged landscapes where a gradient-based approach might converge to the local optima. Such methods have been successfully incorporated for designing reconfigurable organisms \cite{Kriegman.etal2020}, autonomous machines \cite{Lipson.Pollack2000}, and molecular generation for drug discovery \cite{Tripp.Hernandez-Lobato2023}.

Previous work has explored the usage of gradient-free optimization methods for the design of granular materials. Miskin et al. showed that an evolutionary approach can find particle aggregates with the highest/lowest elastic modulus \yrcite{Miskin.Jaeger2013}. In \cite{Parsa.etal2023}, the authors have used a multi-objective evolutionary algorithm to find granular materials that can compute two logic functions at two different frequencies. Therefore, we apply a similar gradient-free optimization method to the design problems explored in this paper to compare their performance to the gradient-based framework proposed here.

In this work, we use the Age-Fitness Pareto Optimization (AFPO) method \cite{Schmidt.Lipson2011}. Evolutionary algorithms generally start with a randomly generated set of candidate solutions (\textit{population}) and at each step of the optimization (\textit{generation}), the best solutions (or non-dominated ones in a multi-objective problem) are selected, slightly modified (with the mutation and cross-over operators) and survived to the next generation. AFPO is a multi-objective evolutionary algorithm that prevents premature convergence and promotes diversity in the solutions by periodically injecting random solutions and allowing newer instances to survive before being dominated by the existing more fitted solutions.

In the experiments in this paper, we employed a direct encoding scheme, defined as a real-valued vector indicating the stiffness values of the particles. A Gaussian mutation operator with a standard deviation of $0.1$ was defined such that it ensures the stiffness remains within the permitted upper and lower bounds. 
In all experiments, a population size of $100$ was used, and each evolutionary trial was conducted for $1000$ generations. $10$ independent runs were performed for each experiment, each starting with a different random initial population with a uniform distribution. The objective functions for each experiment are similar to the loss functions defined for the gradient-based optimization setup and will be introduced in the next section.

\section{Experiments}
\label{sec:results}
To demonstrate the application of our gradient-based design framework we considered three design problems, including an acoustic waveguide, a mechanical AND gate, and a mechanical XOR gate. \cref{tab:simparams} includes the parameter values for the physics model and numerical simulations used in the experiments. The detailed description of the simulation and model parameters is presented in \cref{sec:model}. 

\begin{table}[ht]
\caption{Simulation parameters.}
\label{tab:simparams}
\vskip 0.15in
\begin{center}
\begin{small}
\begin{sc}
\begin{tabular}{lccr}
\toprule
Parameter & Value \\
\midrule
Total Time ($T$)                       & $3 \times 10^3$ \\
Time Step ($\Delta t$)                  & $5 \times 10^{-3}$ \\
Lattice Size ($N = N_x \times N_y$)    & $10 \times 11$\\
Mass ($m$)                              & $1.0$\\
Stiffness $(k)$                         & $\in [1.0, 10.0]$\\
Packing Fraction ($\phi$)               & $0.1$\\
Diameter ($\sigma$)                     & $0.1$ \\
Background Damping ($B$)                & $1.0$\\
Particle-particle Damping ($B_{pp}$)   & $0.0$ \\
Particle-wall Damping ($B_{pw}$)         & $0.0$ \\
\bottomrule
\end{tabular}
\end{sc}
\end{small}
\end{center}
\vskip -0.1in
\end{table}

\subsection{Acoustic Waveguide}
\label{sec:waveguide}
Granular crystals have a discrete band structure with a high \textit{cut-off frequency} that depends on the particle properties (size, Young modulus, and Poisson ratio), boundary conditions, and the applied longitudinal static stress \cite{Franklin.Shattuck2016}. In a harmonically driven system, only waves with frequencies within the pass band can propagate, and waves above the cut-off frequency are attenuated significantly. This phenomenon provides the opportunity to design granular crystals with desired band gaps and tunable filtering behavior that act as acoustic filters and waveguides \cite{Spadoni.Daraio2010}, acoustic switches and logic gates \cite{Li.etal2014a}.
In an acoustic waveguide, the vibrational energy is localized toward specific locations. In the first experiment, we demonstrate how the dynamics of a granular crystal can be tuned by changing the particles' stiffnesses to selectively direct acoustic waves toward one of the two output particles based on the frequency content of the input signal.
\begin{figure}[ht]
\vskip 0.1in
\begin{center}
\centerline{\includegraphics[width=\columnwidth]{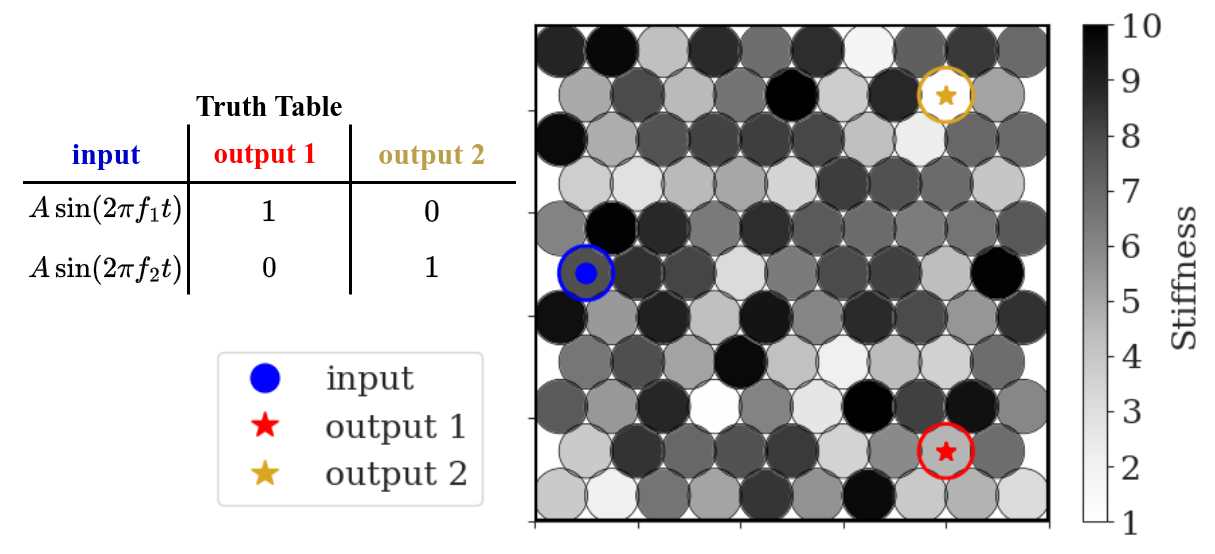}}
\caption{Experimental setup for the acoustic waveguide. The input particle (blue marker) is harmonically vibrated with the amplitude $A$ and one of the two predefined frequencies, $f_1$ or $f_2$. The applied elastic vibrations propagate through the material toward the output ports (red and gold markers). The existence of the input frequency in the displacement signal of the output particles indicates the computational response. Each of the two output particles is expected to only respond to one of the two input frequencies.}
\label{fig:switchSetup}
\end{center}
\vskip -0.1in
\end{figure}
\cref{fig:switchSetup} presents the setup for this experiment. A particle near the left boundary is selected as the input port where the acoustic vibration is injected into the system. The input vibration is in the form of a horizontal sinusoidal wave that displaces particle $i$ from its equilibrium position ($\delta_{0, i}$, see \cref{sec:model}) such that $r^{x}_i(t) = \delta_{0, i} + A \sin{\omega t}$, where $A$ is the amplitude of the input oscillation, and $\omega=2 \pi f$ is its frequency. Similar to the input port, two particles are chosen near the right boundary as the output ports. The horizontal displacements of these output particles are recorded during the simulation, and the wave intensity is calculated as follows:
\begin{equation}
    \label{eq:intensity}
    \hat{Y}_{i} = \sum_{t=\frac{T}{3}}^{T} {(r^{x}_{i}(t))}^2 \quad , i \in {1, 2}
\end{equation}
where $r^{x}_{i}(t)$ is the displacement of the particle in $x$ direction at time $t$, $T$ is the length of the simulation and $i$ indicates the particle index which is $1$ or $2$, representing one of the two output ports.  To remove the effect of transient responses, the first one-third of the simulation time is not included in calculating the wave intensity. The predicted output of the physical neural network is a vector with two scalar values which are the normalized wave intensities at each of the output ports as $\hat{Y}^n=[\frac{\hat{Y}^n_1}{\hat{Y}^n_1 + \hat{Y}^n_2}, \frac{\hat{Y}^n_2}{\hat{Y}^n_1 + \hat{Y}^n_2}]$. $n$ indicates the sample from the training dataset which has two entities and is defined as $D=\{(X^1=[A\sin{2 \pi f_1 t}]_{t=1}^{T}, Y^1=[0, 1]), (X^2=([A\sin{2 \pi f_2 t}]_{t=1}^{T}, Y^2=[1, 0])\}$. To tune the material with a gradient-based optimization framework, we defined a Cross-entropy (CE) loss as follows:
\begin{align}
    & \mathcal{L}_{CE}(\hat{Y}, Y) = -\frac{1}{N} \sum_{n=1}^{N} \log \frac{exp(\hat{Y}^{n}_c)}{exp(\hat{Y}^{n}_1)+exp(\hat{Y}^{n}_2)} \nonumber \\
    & c = argmax(Y^{n}) \label{eq:crossentropy}
\end{align}
where $N$ is the size of the minibatch and $c$ is the port index for the desired output for sample $n$ from the minibatch.

\begin{figure}[h!]
%\vskip 0.1in
\begin{center}
\centerline{\includegraphics[width=\columnwidth]{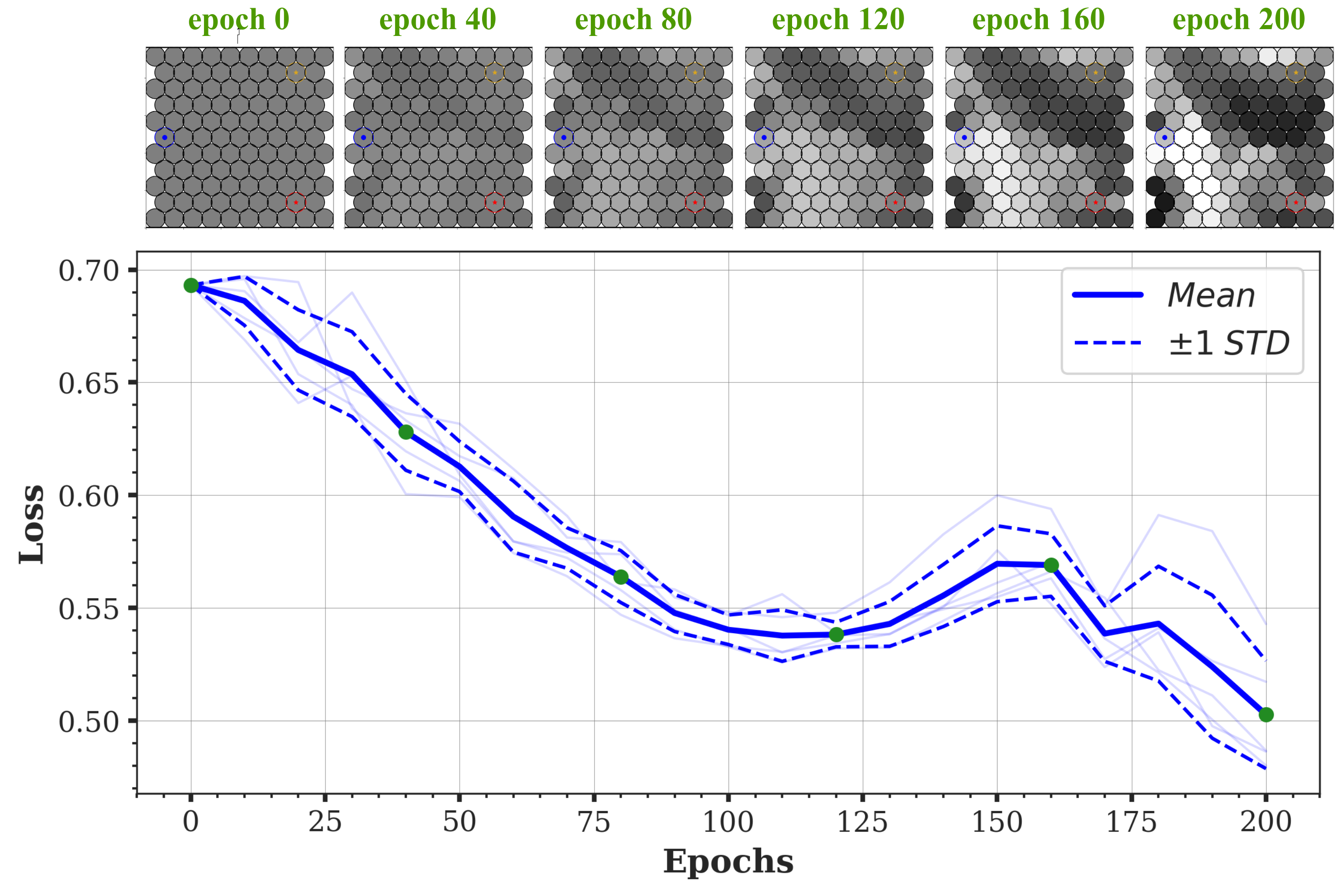}}
\caption{Inverse design of an acoustic waveguide. The training loss is plotted over $200$ epochs. The mean and standard deviation of $5$ independent runs are shown with solid and dashed lines, respectively. Snapshots of the granular crystal are shown at intermediate stages during the training for one example trial.}
\label{fig:switch}
\end{center}
%\vskip -0.1in
\end{figure}

In this experiment, we applied small-amplitude vibrations to enforce the dynamics to stay in the weakly nonlinear regime ($A=10^{-2} \times \sigma = 10^{-3}$). The training dataset $D$ consists of sinusoidal waves at two selected frequencies, $f_1=7[Hz]$, and $f_2=15[Hz]$. As mentioned before, one-hot encoding is utilized to indicate the desired output according to the truth table provided in \cref{fig:switch}.

\begin{figure}[ht]
\vskip 0.1in
\begin{center}
\centerline{\includegraphics[width=\columnwidth]{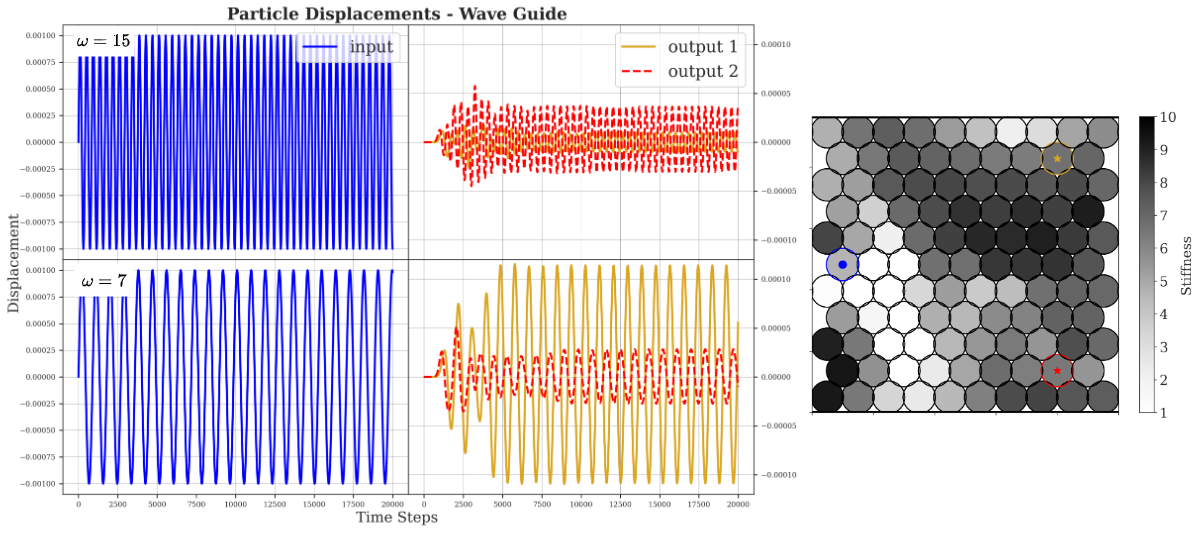}}
\caption{The optimized acoustic waveguide. The stiffness pattern enables the material to direct the vibration toward one of the output ports according to its frequency. The plots on the left show the horizontal displacement of the input (blue) and output particles (red and gold) during the simulation time. The optimized material directs the input vibration toward the top particle when the frequency is $f_1=7[Hz]$ and the bottom particle when the frequency is $f_1=15[Hz]$.}
\label{fig:switchplot}
\end{center}
\vskip -0.1in
\end{figure}

We initialized the stiffness profile ($\theta_0$) with a stiffness value of $k=5.0$ for all the particles at epoch $0$ (see \cref{fig:switch}). Adam optimizer is used with a fixed learning rate of $0.001$ to train the network for $200$ epochs. \cref{fig:switchplot} presents the training loss, averaged over $5$ independent runs.

As it can be seen in the optimized design at epoch $200$ in \cref{fig:switchplot}, the stiffness pattern of the granular crystal is tuned such that the low-frequency vibration is guided toward the top particle. On the other hand, the softer particles around the bottom port enable larger displacements around the second output port when the input is at a high frequency. 

\subsection{Acoustic Logic Gate}
\label{sec:logicgate}
\begin{figure}[ht]
\vskip 0.1in
\begin{center}
\centerline{\includegraphics[width=\columnwidth]{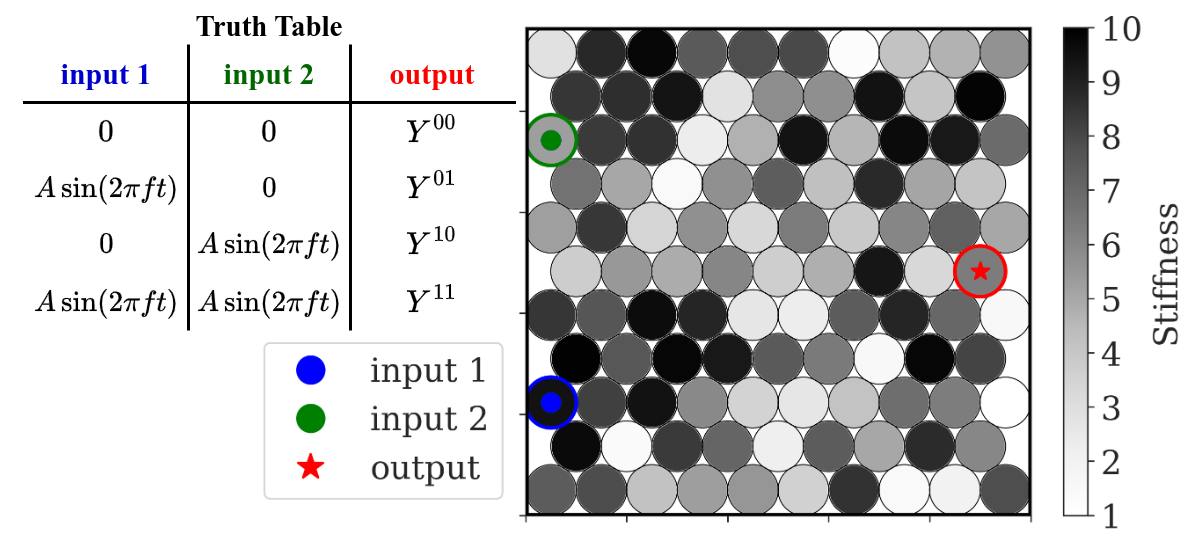}}
\caption{Experimental setup for the acoustic logic gates. The granular crystal is made of circular particles with various stiffness values represented in different shades of grey. The two particles on the left (with green and blue markers) are selected as the input ports where the external sinusoidal oscillations are applied to the system. As noted in the truth table, the $0$/$1$ values at the ports are encoded as harmonic vibrations at a particular frequency $f$. The \textbf{00} case has a trivial solution that is always correct: since there are no power sources in the system, the lack of input vibrations means the output is $0$, which is true for both AND and XOR gates. For the other three cases (\textbf{01}, \textbf{10}, and \textbf{11}) the horizontal displacement of the output particle from its equilibrium position ($r^x(t)$) is measured to infer the output of the logic function.}
\label{fig:logicgate}
\end{center}
\vskip -0.1in
\end{figure}

To demonstrate the computational capabilities of new physical substrates as alternatives to traditional digital electronic devices, many studies show designs for basic logic gates as a reasonable benchmark \cite{Yasuda.etal2021}. In this paper, we first showed the design of an acoustic AND gate. To showcase the exploitation of the nonlinear dynamics of granular crystals for mechanical computing, we also investigated the design of an XOR gate as it performs a nonlinear input-output transformation. \cref{fig:logicgate} shows our experimental setup for the realization of acoustic logic gates.

The input and output signals are mechanical vibrations of the selected particles in the granular crystals. To design logic functions, we first need to define a representation relation that dictates how we encode the binary values. Here, we measure the horizontal displacement of the particles from their equilibrium positions: using the amplitude of the input vibration as the baseline, a significant periodic displacement is interpreted as the binary \textbf{$1$} value, and a negligible one is interpreted as \textbf{$0$}. We apply sinusoidal vibrations as the input and for the experiments in this section, we fixed the operational frequency of the logic gate at a predetermined frequency ($f=15[Hz]$), which was chosen according to the material properties of the granular crystal and its frequency spectrum. The amplitude of oscillations ($A$) is $10^{-3}$, which is the same as the experiments in \cref{sec:waveguide}.

To tune the particles' stiffness values, we define the \textit{L-1} loss function (Mean Absolute Error, MAE) between the intensity of the horizontal displacement of the output particle ($\hat{Y}=[r^x_{output}(t)]_{t=\frac{2T}{3}}^T$) and the desired output ($Y$) as follows:
\begin{align}
    \mathcal{L}_{MAE}(Y, \hat{Y}) & = \parallel Y-\hat{Y} \parallel_{MAE} \nonumber \\
    & = \frac{1}{N} \sum_{n=1}^{N}|{(Y^n-\hat{Y}^n}| \label{eq:MAEloss}
\end{align}
where $N$ is the number of samples in the training dataset $D$, and the superscript $n$ represents the output for each sample. When calculating the wave intensity, we only use the last one-third of the simulation time ($[\frac{2T}{3}, T]$, where $T$ indicates the total simulation) to ignore the transient part of the signals. In the following sections, we provide our results for designing two basic logic gates, an AND gate, and an XOR gate.

\subsubsection{AND Gate}
\label{sec:andgate}
We start with designing an AND gate because, due to its linear nature, we expect the design process to be straightforward. Although, due to the strong nonlinearity in the system, it is theoretically capable of more complex computations, the high-dimensional parameter space ($10 \times 11$ real numbers in $[1.0, 10.0]$) can make the gradient-based optimization challenging. The training dataset $D$ is made of $3$ time series for the three cases defined in \cref{fig:logicgate} as follows:
\begin{align}   
    & D=\{ \nonumber \\
    & (X^{01}=[X_1=A\sin{2 \pi f t}, X_2=0]_{t=1}^{T}, Y^{01}=[0]_{t=1}^{T}), \nonumber \\
    & (X^{10}=[X_1=0, X_2=A\sin{2 \pi f t}]_{t=1}^{T}, Y^{10}=[0]_{t=1}^{T}), \nonumber \\
    & (X^{11}=[X_1=A\sin{2 \pi f t}, X_2=A\sin{2 \pi f t}]_{t=1}^{T}, \nonumber \\
    & Y^{11}=[A\sin{2 \pi f t}]_{t=1}^{T})\}  \label{eq:ANDdataset}
\end{align}

\begin{figure}[t]
\vskip 0.1in
\begin{center}
\centerline{\includegraphics[width=\columnwidth]{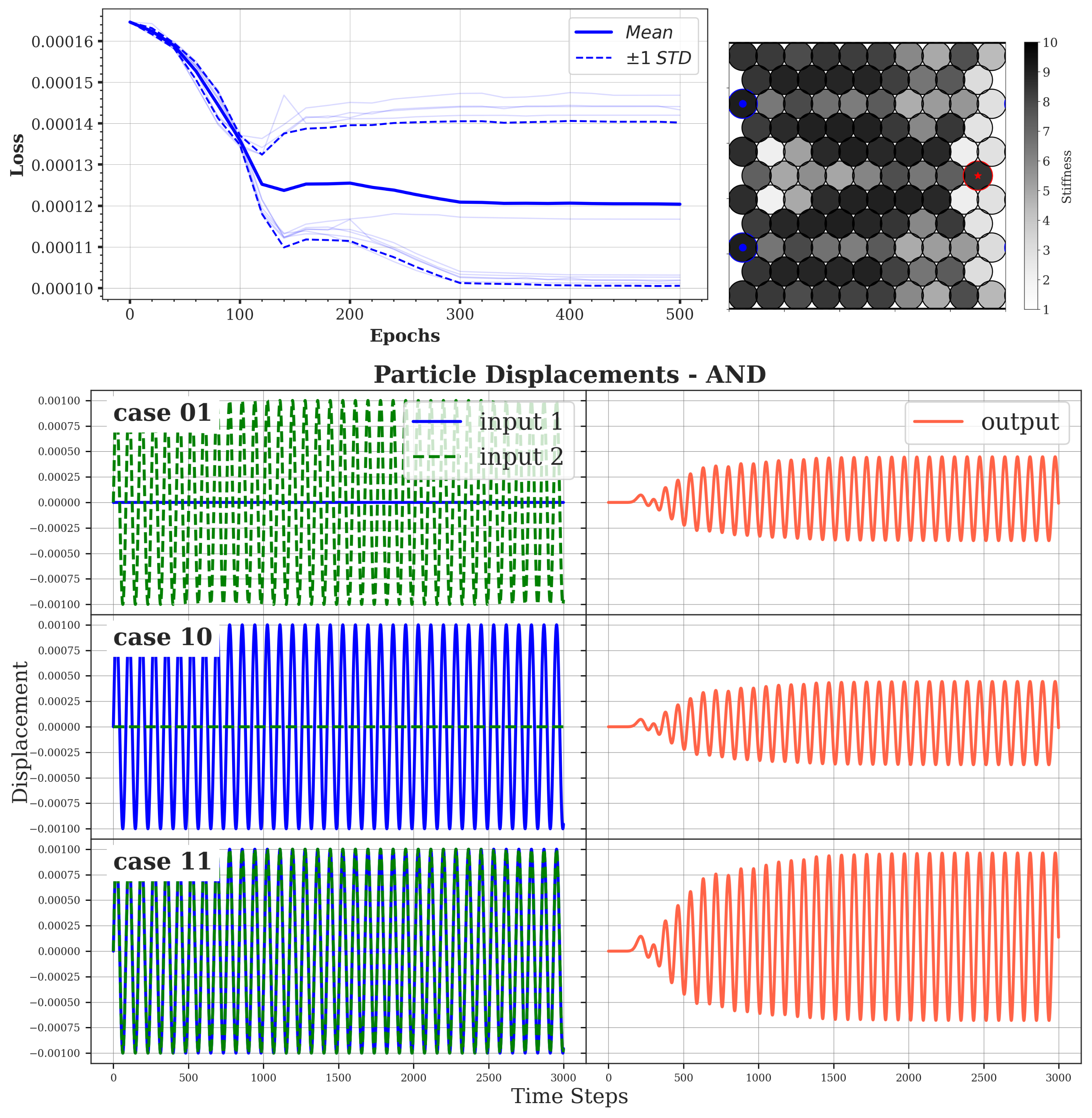}}
\caption{Gradient-based design of an acoustic AND gate with fixed-value initialization. The granular crystal is initialized as a homogeneous assembly of particles with a stiffness value located in the middle of the permitted range. Each light graph in the top left panel shows the training loss for one of the $10$ independent trials over $500$ epochs. One example of the optimized material is shown on the right. The plots on the bottom show the inputs and output of the logic gate, as the horizontal displacement of the particles in time. Each row shows one of the three cases.}
\label{fig:and_73}
\end{center}
\vskip -0.1in
\end{figure}

\begin{figure}[t]
\vskip 0.1in
\begin{center}
\centerline{\includegraphics[width=\columnwidth]{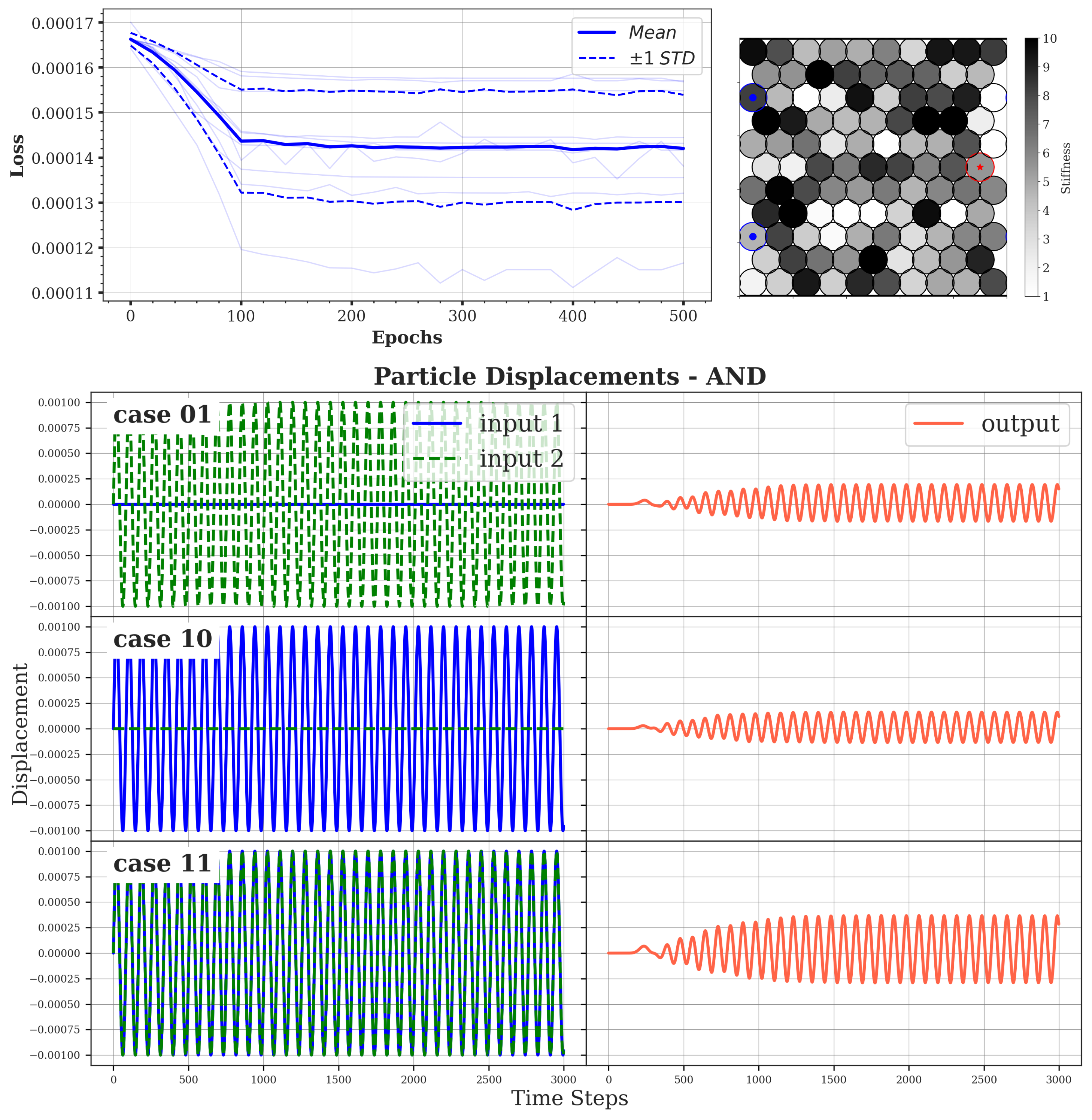}}
\caption{Gradient-based design of an acoustic AND gate with random initialization. The granular crystal is initialized with random stiffness values drawn uniformly from the permitted range ($\in [1.0, 10.0]$). The light blue graphs in the top left panel show the training loss of the $10$ independent trials. One of the optimized materials is shown on the right. The plots on the bottom show the inputs and output of the logic gate in each of the three cases \textbf{01}, \textbf{10}, and \textbf{11}.}
\label{fig:and_80}
\end{center}
\vskip -0.1in
\end{figure}

We used the same simulation parameters as reported in \cref{tab:simparams}. In our preliminary investigations, we noticed that the initial values of the trainable parameters ($\theta$) affect the performance of the optimization when all other simulator and optimization parameters are fixed. Therefore, we conducted two experiments, one starting with a homogeneous configuration of identical particles with a stiffness value of $k=0.5$ and the other with randomly initialized values in the range $[1.0, 10.0]$. \cref{fig:and_73} and \cref{fig:and_80} present the training loss and an example of the optimized material from one of the $10$ independent trials in each of the two setups. It should be noted that the sudden changes in the training loss at specific instances are due to the incorporation of an adaptive learning rate. We incorporated a multi-step adaptive learning rate with a decay rate of $\gamma=0.1$ and step sizes at $[150, 300, 400]$ epochs for the fixed-value initialization and $[100, 200, 300]$ epochs for the randomly initialized case. The starting value of the learning rate is set to $lr=0.001$ in both cases.

\subsubsection{XOR Gate}
\label{sec:xorgate}
\begin{figure}[t]
\vskip 0.1in
\begin{center}
\centerline{\includegraphics[width=\columnwidth]{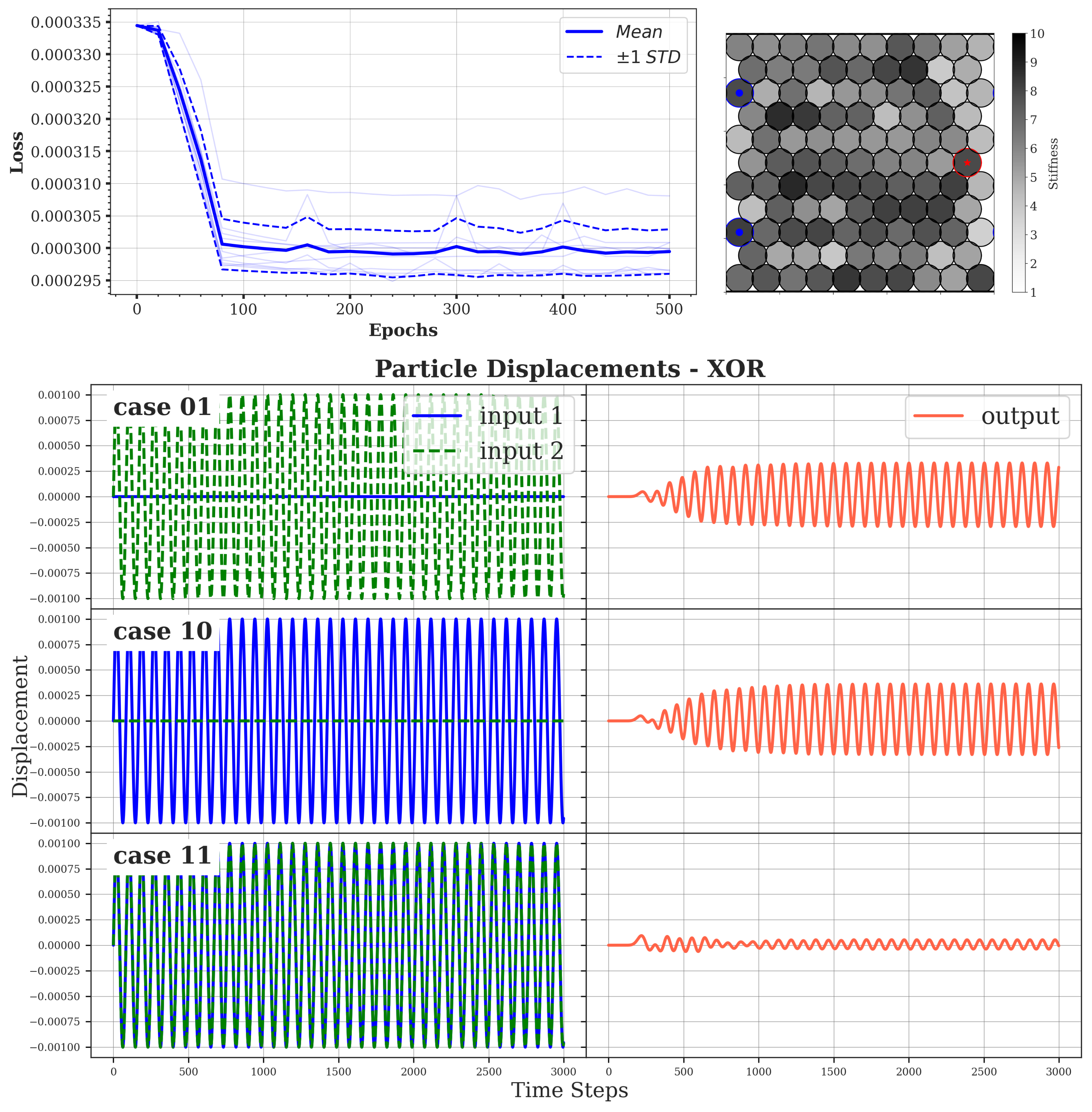}}
\caption{Gradient-based design of an acoustic XOR gate with fixed-value initialization. The granular crystal is initialized as a homogeneous assembly of particles with a stiffness value located in the middle of the permitted range. Each light graph in the top left panel shows the training loss for one of the $10$ independent trials over $500$ epochs. One example of the optimized material is shown on the right. The plots on the bottom show that, as we expect, the output particle oscillates with a higher amplitude when only one of the input ports is vibrated. This is consistent with the desired functionality of an XOR gate.}
\label{fig:xor_73}
\end{center}
\vskip -0.1in
\end{figure}

\begin{figure}[t]
\vskip 0.1in
\begin{center}
\centerline{\includegraphics[width=\columnwidth]{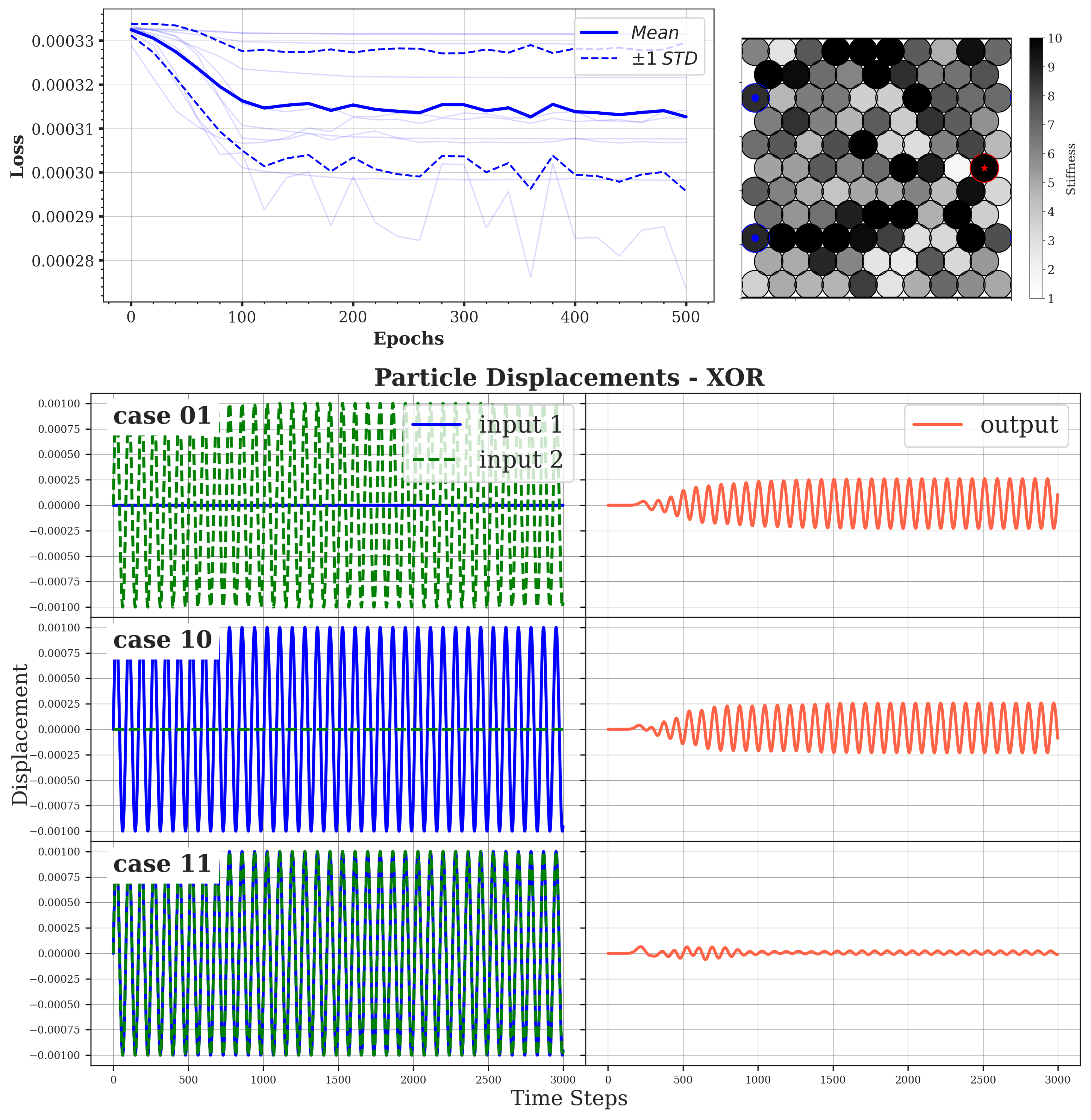}}
\caption{Gradient-based design of an acoustic XOR gate with random initialization. The granular crystal is initialized with random stiffness values drawn uniformly from the permitted range ($[1.0, 10.0]$). The light blue graphs in the top left panel show the training loss of the $10$ independent trials. One of the optimized materials is shown on the right. The plots on the bottom show the inputs and output of the logic gate in each of the three cases \textbf{01}, \textbf{10}, and \textbf{11}.}
\label{fig:xor_80}
\end{center}
\vskip -0.1in
\end{figure}

We repeated the design problem for an XOR gate with the same set of parameters for the simulator and the optimizer. As in the previous section, a dataset containing the time series of the inputs and the target is produced and incorporated for optimizing the stiffness values of the particles as follows:
\begin{align}   
    & D=\{ \nonumber \\
    & (X^{01}=[X_1=A\sin{2 \pi f t}, X_2=0]_{t=1}^{T}, \nonumber\\
    & Y^{01}=[A\sin{2 \pi f t}]_{t=1}^{T}), \nonumber \\
    & (X^{10}=[X_1=0, X_2=A\sin{2 \pi f t}]_{t=1}^{T}, \nonumber \\
    & Y^{10}=[A\sin{2 \pi f t}]_{t=1}^{T}), \nonumber \\
    & (X^{11}=[X_1=A\sin{2 \pi f t}, X_2=A\sin{2 \pi f t}]_{t=1}^{T}, \nonumber \\
    & Y^{11}=[0]_{t=1}^{T})\}  \label{eq:XORdataset}
\end{align}
The optimization results for the two initial conditions, fixed-value and random, are shown in \cref{fig:xor_73} and \cref{fig:xor_80}.  

\section{Discussions}
\label{sec:discussion}
The results presented in the previous section show that our gradient-based optimization framework can find granular crystals with the desired dynamic wave responses. Comparing the two experiments with different initialization of the stiffness values suggests that starting with a homogeneous stiffness pattern promotes some degree of symmetry in the $y$ direction which is expected because the input ports are placed at the same distance from the output port (see \cref{fig:logicgate}) and the logic function is inherently symmetric with respect to the two inputs.

As mentioned in \cref{sec:methods}, gradient-free optimization methods have been incorporated previously in similar design problems for computational granular materials \cite{Parsa.etal2023}. To compare the efficiency and performance of our proposed gradient-based framework, we applied a gradient-free optimization method (\cref{sec:EA}) to the logic gate design problems discussed above. The results of this experiment are provided in \cref{sec:EAresults}. Evaluating the best solutions from this optimization method (\cref{fig:EA_t} and \cref{fig:EA_f}) shows that they don't exhibit the desired response in all three cases. Since we have three different objectives combined into one loss value (\cref{eq:metric}), the desired solution will offer a tradeoff between them. However, the gradient-free optimization method finds solutions that perform well in some cases while ignoring the others. For example, for the AND gate in \cref{fig:EA_t}, the material produces the desired output in the $01$ and $10$ cases, but not in the $11$ case.

We can compare the computing resources needed for each optimization method by the number of the simulations of granular crystal needed during the optimization. In the gradient-based approach, each epoch in the logic gate experiment (\cref{sec:logicgate}) consists of $3$ evaluations, so the optimization performs $1.5e2$ simulations after $500$ epochs to find the best design. However, the population-based methods keep a set of $100$ solutions at each step (population size=$100$), therefore the optimization has performed $300$ simulations of the physics model in each step and the total number of simulations at the end of $10^3$ optimization steps is $3e5$. Thus, the material design space is explored efficiently with the gradient-based optimization method. However, the gradient-based method needs to retain the computational graph for backpropagation of gradients of loss, therefore the memory requirements are higher than the gradient-free methods.

\begin{figure*}[ht]
\vskip 0.1in
\begin{center}
\centerline{\includegraphics[width=\linewidth]{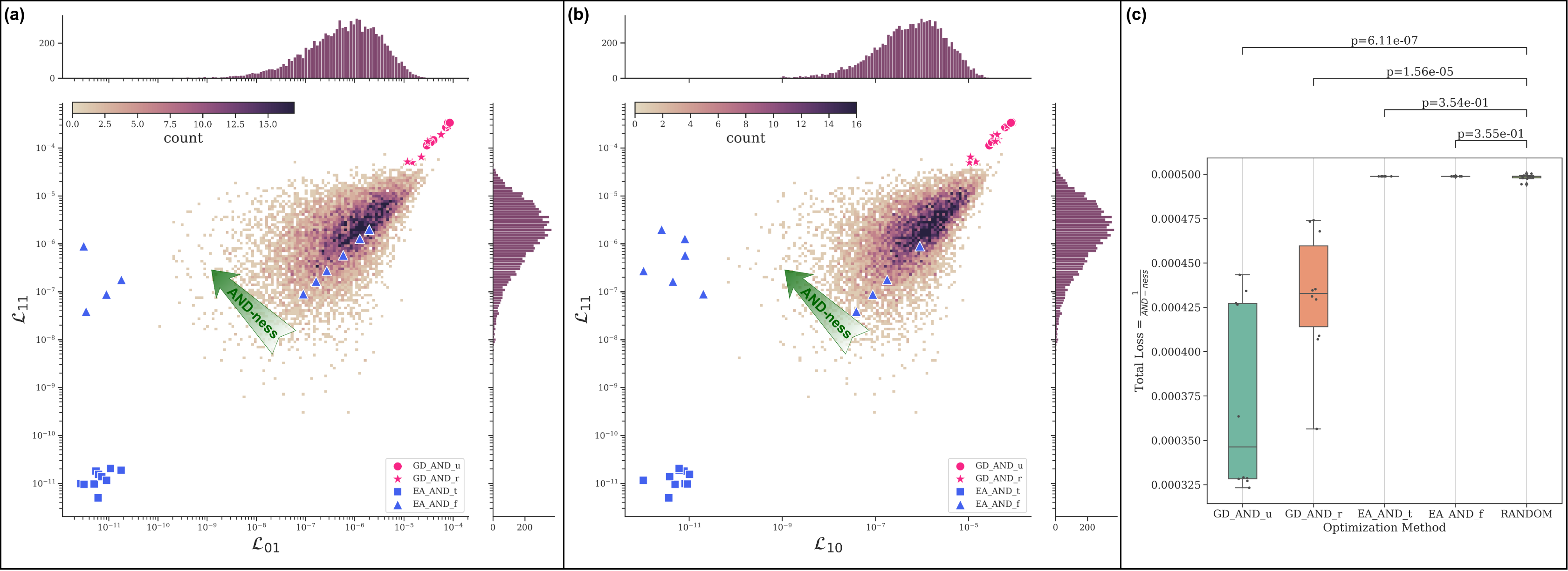}}
\caption{Exploration of the design space of the acoustic AND gates. (a), (b): the distribution of ${10}^4$ random configurations in the space of partial loss values (\cref{eq:metric}). The green arrow indicates the direction with increasing ``AND-ness'' in the material. The blue markers (\textit{EA}) represent the solutions from the gradient-free optimization method (\cref{sec:EAresults}) with the loss defined in the time domain (\textit{EA\_AND\_t}, \cref{fig:EA_t}) and frequency domain (\textit{EA\_AND\_f}, \cref{fig:EA_f}). The pink markers (\textit{GD}) show the solutions from the gradient-based optimization experiments (\cref{sec:GD}) in each of the two initialization cases, random (\textit{GD\_AND\_r}, \cref{fig:and_73}) and fixed-value initialization (\textit{GD\_AND\_u}, \cref{fig:and_80}). (b): boxplots showing the distribution of the best solution from $10$ independent optimization runs and $10$ randomly generated configurations. The vertical axis is the \textbf{Total Loss} ($\mathcal{L}_{Total}$) that is minimized during the optimization to maximize the \textbf{``AND-ness''} in the material. The p-values are shown on the top.}
\label{fig:EAvsGD_and}
\end{center}
\vskip -0.1in
\end{figure*}

\begin{figure*}[ht]
\vskip 0.1in
\begin{center}
\centerline{\includegraphics[width=\linewidth]{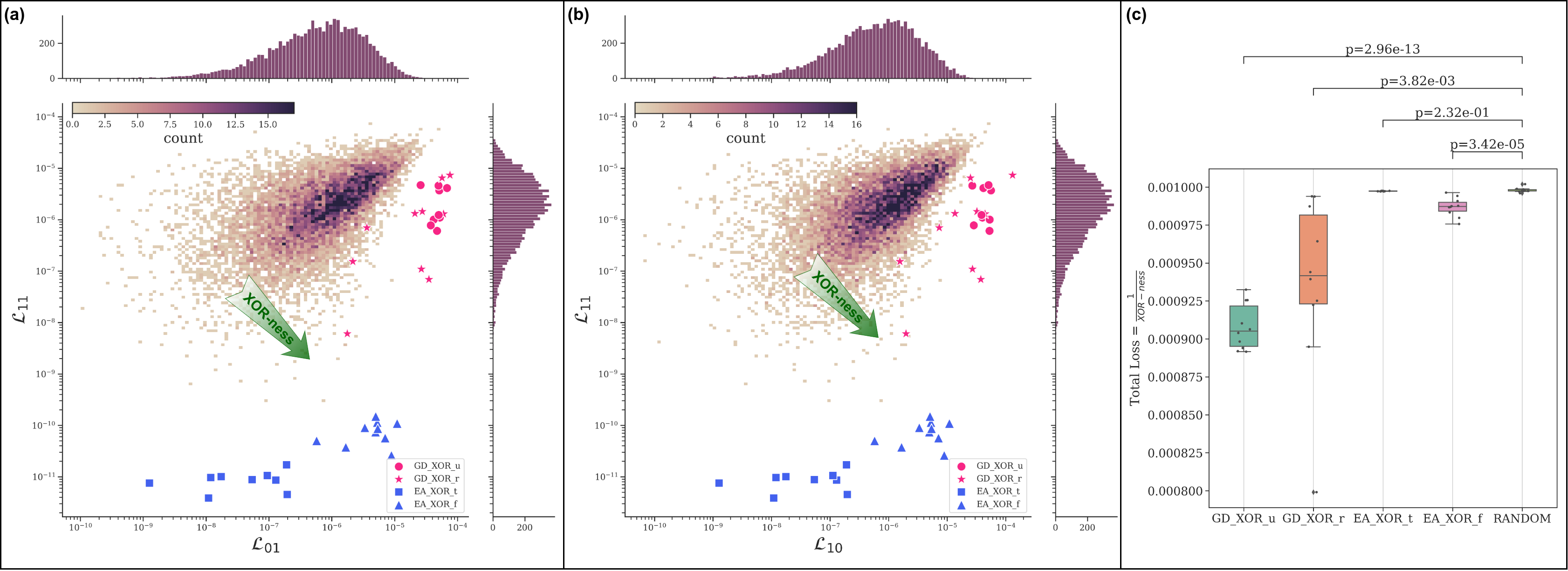}}
\caption{Exploration of the design space of the acoustic XOR gates. (a), (b): the distribution of ${10}^4$ random configurations in the space of partial loss values (\cref{eq:metric}). The green arrow indicates the direction with increasing ``XOR-ness'' in the material. The blue markers (\textit{EA}) represent the solutions from the gradient-free optimization method (\cref{sec:EAresults}) with the loss defined in the time domain (\textit{EA\_XOR\_t}, \cref{fig:EA_t}) and frequency domain (\textit{EA\_XOR\_f}, \cref{fig:EA_f}). The pink markers (\textit{GD}) show the solutions from the gradient-based optimization experiments (\cref{sec:GD}) in each of the two initialization cases, random (\textit{GD\_XOR\_r}, \cref{fig:xor_73}) and fixed-value initialization (\textit{GD\_XOR\_u}, \cref{fig:xor_80}). (b): boxplots showing the distribution of the best solution from $10$ independent optimization runs and $10$ randomly generated configurations. The vertical axis is the \textbf{Total Loss} ($\mathcal{L}_{Total}$) that is minimized during the optimization to maximize the \textbf{``XOR-ness''} in the material. The p-values are shown on the top.}
\label{fig:EAvsGD_xor}
\end{center}
\vskip -0.1in
\end{figure*}

To gain an understanding of the shape of the design landscape shape and the complexity of the optimization problem, we performed a random search by generating ${10}^4$ random configurations and evaluating them using the loss function defined in \cref{eq:metric}. The distributions of the random configurations along with the optimized solutions are shown in the space of the three loss values in \cref{fig:EAvsGD_and} and \cref{fig:EAvsGD_xor}.
\begin{align}
    & \mathcal{L}_{Total} = \mathcal{L}_{01} + \mathcal{L}_{10} + \mathcal{L}_{11} \nonumber \\
    & \mathcal{L}_{01} = \parallel Y^{01}-\hat{Y}^{01} \parallel_{MAE} \nonumber \\
    & \mathcal{L}_{10} = \parallel Y^{10}-\hat{Y}^{10} \parallel_{MAE} \nonumber \\
    & \mathcal{L}_{11} = \parallel Y^{11}-\hat{Y}^{11} \parallel_{MAE} \label{eq:metric}
\end{align}
Using the total loss defined above, we computed p-values to indicate the significance of the optimized configurations compared to the random ones. We can see that the gradient-based method has found better designs than the random search and gradient-free method for both problems. The distribution of the solutions from the gradient-free method shows that this approach has prioritized some of the loss terms over the others and thus is not able to find solutions that are significantly better than the random search.

\section{Conclusion}
\label{sec:conclusions}
Motivated by the growing interest in unconventional computing substrates, in this paper, we explored the application of gradient-based optimization frameworks for the design of computational granular crystals. We showed that by developing a differentiable simulator, we can employ gradient-based optimization methods to tune the material properties of the constituent particles of a granular crystal to allow for the desired wave responses.

Unlike previous work such as Mechanical Neural Networks that train the physical system directly \cite{Lee.etal2022a}, here we train the physics model in a differentiable simulator. Therefore, transferring the designs to reality can be challenging because of the discrepancies between the real and simulated systems. In future work, we can address this by including random amounts of reasonable noise to the parameters and finding designs that tolerate reasonable manufacturing errors. Despite this, our approach can still provide valuable insights into the design space of granular crystals. For example, using our design framework we can investigate which physical properties of the physical substrate offer more opportunities for a desired computational task.

Similar to traditional RNNs, the size of the hidden state of the network (the number of particles in the granular crystal) determines its memory capacity and computational complexity. Previous work has concluded that gradient-free optimization methods are not scalable and cannot search the parameter space of such large models effectively. Our results prove the power of gradient-based optimization in this problem but such methods can fail in non-convex landscapes. Moreover, unlike population-based techniques, our framework cannot find a set of Pareto-optimal solutions to problems with multiple objectives. Designing multifunctional materials has attracted immense attention in recent years and inverse design methods are needed that can find solutions with various trade-offs between multiple objectives and constraints. Therefore, it is possible that augmenting the gradient-free methods with gradient-based approaches facilitate the design of such multifunctional granular machines in the future.

\section*{Acknowledgements}
We would like to acknowledge financial support from the National Science Foundation under the DMREF program (award number: $2118810$).

Computations were performed, in part, on the Vermont Advanced Computing Center.

\bibliography{refs}
\bibliographystyle{style}

\newpage
\appendix
\onecolumn
\section{Appendix}
\subsection{Physics Model}
\label{sec:model}
The granular crystals discussed in this paper are finite-length two-dimensional configurations of spherical particles with identical diameters and variable elasticity placed on a horizontal flat surface (\cref{fig:model}). The system is a macroscopic scale granular system (particle sizes are in the millimeter-to-centimeter range), so the only forces acting on each particle are the finite-range repulsive interparticle contact forces. On the scale of particle contacts, we consider normal forces resulting from the adjacent particles' overlaps and ignore the tangential forces and particle rotations. With these assumptions, the local potential between each pair of particles $i$ and $j$ can be written as:

\begin{equation}
    \label{eq:potential}
    V_{ij}(r_{ij}) = \frac{\epsilon}{\alpha}{(1-\frac{r_{ij}}{\sigma_{ij}})}^\alpha \Theta(1 - \frac{r_{ij}}{\sigma_{ij}})
\end{equation}

where $\epsilon$ is the characteristic energy scale, $r_{ij}$ is the particles' separation, and $\sigma_{ij}$ is the center-to-center separation at which the particles are in contact without any deformation. In the case of spherical particles with diameters $\sigma_i$ and $\sigma_i$ we'll have: $\sigma_{ij}=\frac{\sigma_i+\sigma_j}{2}$. $\Theta$ in this equation is the Heaviside function, which ensures that the potential field is one-sided, meaning that the particles only affect their adjacent neighbors when they are overlapping:

\begin{equation}
    \Theta(1 - \frac{r_{ij}}{\sigma_{ij}}) = 
        \left\{ \begin{aligned}
            & 0 & {\frac{r_{ij}}{\sigma_{ij}} \geq 1} \\
            & 1 & {\frac{r_{ij}}{\sigma_{ij}} < 1}
        \end{aligned} \right.
\end{equation}

This is the simplest model for a granular crystal that neglects special aspects such as particles' rotation and alignment, which might be more important in higher dimensional experimental setups but are negligible in smaller scales.
The separation between two spherical particles is computed based on their Cartesian coordinates as follows:

\begin{equation}
    |{r}_{ij}| = |\overrightarrow{r_i}-\overrightarrow{r_j}| = \sqrt{x_{ij}^2 + y_{ij}^2} 
\end{equation}

In \cref{eq:potential}, $\alpha$ determines the nonlinearity of the contact force. In this paper, we consider %two cases: linear ($\alpha=2$) and 
Hertzian ($\alpha=\frac{5}{2}$) contacts to provide enough nonlinearity in the physical substrate for performing the desired computations. Interparticle forces can be obtained by taking the derivative of the potential ($V_{ij}$) with respect to the displacement:

\begin{equation}
    \label{eq:force}
    \begin{aligned}
    F_{ij} & = -\frac{\partial V_{ij}(r_{ij})}{\partial r_{ij}} \\
    & = \frac{\epsilon}{\sigma_{ij}}{(1-\frac{r_{ij}}{\sigma_{ij}})}^{\alpha-1} \Theta(1 - \frac{r_{ij}}{\sigma_{ij}}) \frac{\partial r_{ij}}{\partial (\text{$x_{ij}$ or $y_{ij}$})}
    \end{aligned}
\end{equation}

We assume that the particles have similar mass ($m$) but can have different stiffness values. In this case, $\epsilon$ can be calculated using the effective stiffness as follows:

\begin{equation}
    \epsilon_{ij} = 
    \left\{ \begin{aligned}
        k_i = k_j: & \text{ if $k_i=k_j$} \\
        \frac{k_i \times k_j}{k_i + k_j}: & \text{ if $k_i \neq k_j$} 
    \end{aligned}\right.
\end{equation}

Using the above notation, we can write Newton's equations of motion as:

\begin{equation}
    \label{eq:motion}
    m_{i} \ddot{r}_{i} = F_{i} = \sum^N_{j = 1, j \neq i} F_{ij} + F_{ext}
\end{equation}

where the first term is the total force from the neighboring particles, and the second term is the system's external forces, which include the interaction force from the walls (in case of a fixed boundary condition) and the excitation applied to the system in the form of harmonic vibrations. Using \cref{eq:force}, we can obtain the partial forces in a one-dimensional system as:

\begin{align}
        & F^x(r_{ij}) = \frac{\epsilon_{ij}}{\sigma_{ij}} {(1-\frac{x_{ij}}{\sigma_{ij}})} ^{\alpha-1} \quad \Theta(1-\frac{x_{ij}}{\sigma_{ij}})
        \nonumber \\
        & F^x_{iw} = \frac{\epsilon}{\sigma_{i}/2} {(1-\frac{x_{i}-x_{w}}{\sigma_{i}/2})} ^{\alpha-1} \quad \Theta(1-\frac{x_{i}-x_{w}}{\sigma_{i}})    
\end{align}

where $F^x_{iw}$ is the force between particle $i$ and the wall placed at $x_w$. In a two-dimensional system, the forces are given by:

\begin{align}
        & F^x(r_{ij}) = \frac{\epsilon_{ij}}{\sigma_{ij}} {(1-\frac{r_{ij}}{\sigma_{ij}})} ^{\alpha-1} \quad \frac{x_{ij}}{r_{ij}} \quad \Theta(1-\frac{r_{ij}}{\sigma_{ij}})
        \nonumber \\
        & F^y(r_{ij}) = \frac{\epsilon_{ij}}{\sigma_{ij}} {(1-\frac{r_{ij}}{\sigma_{ij}})} ^{\alpha-1} \quad \frac{y_{ij}}{r_{ij}} \quad \Theta(1-\frac{r_{ij}}{\sigma_{ij}})
        \nonumber \\
        & F^x_{iw} = \frac{\epsilon}{\sigma_{i}/2} {(1-\frac{x_{i}-x_{w}}{\sigma_{i}/2})} ^{\alpha-1} \quad \Theta(1-\frac{x_{i}-x_{w}}{\sigma_{i}})
        \nonumber \\
        & F^y_{iw} = \frac{\epsilon}{\sigma_{i}/2} {(1-\frac{y_{i}-y_{w}}{\sigma_{i}/2})} ^{\alpha-1} \quad \Theta(1-\frac{y_{i}-y_{w}}{\sigma_{i}})
\end{align}

\subsubsection{Dissipation}
To capture the dissipation effects in real granular crystals, we remove the Hamiltonian assumption and incorporate a dash-pot form of dissipation with a velocity-dependent functional form and characteristic constants for background, particle-particle, and particle-wall interactions. This adds extra terms to the equations of motion of each particle $i$ (\cref{eq:motion}), and we'll have:

\begin{equation}
    \label{eq:motionDamping}
    m_{i} \ddot{r}_{i} = F_{i} = \sum^N_{j = 1, j \neq i} F_{ij} + \sum_{walls} F_{iw} - F_{ib} + F_{ext}
\end{equation}

where $m_i$ is the mass of particle $i$ and the dissipation $F_{ib}$ is:
\begin{align}
    F_{ib}  &= B v_i \nonumber \\
             &+ \sum_{j} B_{pp} v_{ij} \Theta(1-\frac{r_{ij}}{\sigma_{ij}}) \nonumber \\
             &+ \sum_{walls} B_{pw} v_{i} \Theta(1-\frac{r_{i}-r_{w}}{\sigma_{i}}), \nonumber \\
             & \quad v_i=\frac{\partial r_{i}}{\partial t}, \quad v_{ij}=v_i-v_j
\end{align}

The damping coefficients ($B$: background damping, $B_{pp}$ particle-particle damping, and $B_{pw}$ particle-wall damping) are usually determined by curve fitting in an experimental setup.

\subsubsection{Strength of Nonlinearity}
The discrete nature of the granular crystals makes it possible to tune the degree of nonlinearity in the system's dynamics. As was mentioned in the introduction, we study the particles in a confined space and under an initial static compression, which keeps the particles in place. By controlling the amount of precompression, the system can transition from a linear to a strongly nonlinear regime. To study this effect, we introduce the packing fraction as follows: 

\begin{equation}
    \phi = \frac{A_{part}}{A_{sys}}
\end{equation}

where $A_{part}$ is the sum of the area of the particles, and $A_{sys}$ is the area of the system (the confining box).
In this paper, we assume that the unforced system is mechanically stable and above the jamming state ($\phi_{j}$, mechanical rigidity). Because of the initial precompression, the system will have a nonzero initial energy that depends on the amount of overlap between adjacent particles. To find the initial particle positions ($\delta_{0, i}$) and interparticle overlaps, an energy minimization algorithm is used, which will be explained in the next subsection.

\subsubsection{Numerical Simulation}
\label{sec:simulation}
We use the Discrete Element Method (DEM) \cite{Cundall.Strack1979} to simulate the motion of the interacting particles in a granular crystal. The simulation starts with the initial configuration and updates the positions and velocities by numerically integrating the equations of motion (\cref{eq:motion}).  
Since our granular packings are made of particles with various material properties and are initially compressed with a uniform force, we need to ensure that the initial configuration is statically stable (the ground state $u=0$ and $\dot{u}=0$ is the minimum of energy). Here, we adopt a packing generation protocol that applies successive compression/decompression by changing the particle sizes \cite{Gao.etal2009a, Franklin.Shattuck2016}. An energy minimization technique, Fast Inertial Relaxation Engine (FIRE) \cite{EcheverriRestrepo.etal2013}, is used to relax the interparticle forces and reach equilibrium following the repetitive deformations. With this method, we can find the particles' initial positions for a mechanically stable configuration with a given boundary condition \cite{Asenjo-Andrews2013}. 
To integrate the equations of motion, we use the Velocity Verlet integration algorithm, which is derived by the Taylor expansion of the particle positions at a small period $\Delta t$ around time $t$ \cite{Verlet1967}.

\section{Spectral Loss Function}
\label{sec:spectralLoss}
Over the last decade, research in unconventional computing paradigms has progressed to a great extent. Many natural and artificial substrates have been proposed to be capable of some degree of computation. Therefore, assessing the computational capacity of various substrates and their intrinsic properties for expressive computation is of great importance.

\begin{figure}[h!]
\vskip 0.1in
\begin{center}
\centerline{\includegraphics[width=0.85\linewidth]{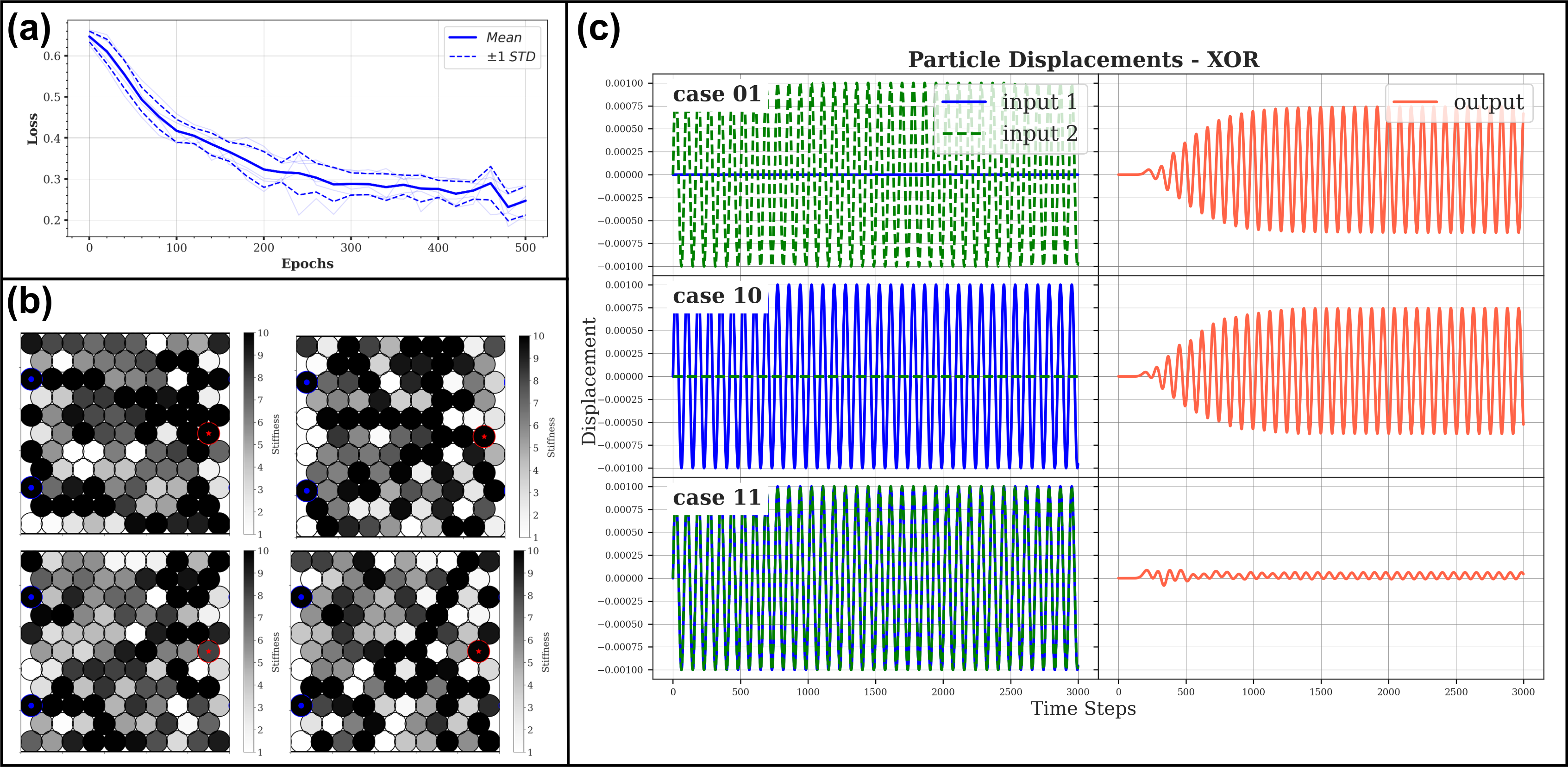}}
\caption{Gradient-based design of an acoustic XOR gate in a granular crystal with the spectral loss function \cref{eq:gain}. The particles' stiffness values are initialized randomly in the permitted range $[1.0, 10.0]$. (a): plot showing the training loss versus epochs for $10$ independent runs, with the average and standard deviation shown in solid and dashed blue lines, respectively. (b): four best configurations from four different trials. (b): evaluation of the designed logic gate in time domain. The displacement of the input particles (green and blue) and output particle (red) is shown in each of the three cases ($`01`, `10`, `11`$) for one of the best solutions.}
\label{fig:xorfft}
\end{center}
\vskip -0.1in
\end{figure}

One of the key advantages of granular materials is their ability to exhibit different dynamical responses depending on the frequency of the propagating waves. This property has been exploited for \textit{polycomputing} in granular materials in recent studies  \cite{Bongard.Levin2023, Parsa.etal2023}. So, defining the loss function in the frequency domain can enable us to exploit the computational power of granular crystals to the full extent. To address this, we revised the loss function introduced in \cref{sec:methods} by incorporating the Fast Fourier Transform (FFT) of the displacement signals. Since the encoding of the logic gates is based on the harmonic vibrations at predefined frequencies ($\omega$), we calculate the systems' gain ($G^{n}$) in each of the three input cases as follows:
\begin{equation}\label{eq:gain}
G^{n}(\omega) = \frac{|\hat{f}(Y^{n})|_{\omega}}{|\hat{f}(X^n)|_{\omega}}, \quad n \in \{01, 10, 11\}
\end{equation}
where $|\hat{f}(Y^{n})|_{\omega}$ indicates the magnitude of the FFT of the output particle's displacement in $x$ direction (as defined in \cref{fig:logicgate}) at the driving frequency $\omega=2\pi f$, and $|\hat{f}(X^n)|_{\omega}= |\hat{f}(X^n_1)|_{\omega} + |\hat{f}(X^n_2)|_{\omega}$ is the same quantity calculated for the first and second input ports. We used \cref{eq:gain} to evaluate each granular assembly and calculate the loss in the frequency domain. In order to obtain a better estimation of the gain, we removed the transient part of the displacement signal and only used the last ${10}^3$ time steps for computing the FFT. \cref{fig:xorfft} shows the results of the optimization with this setup.

Comparing the functionality of the designed logic gate through the time-domain plots provided in \cref{fig:xorfft} with the plots in \cref{fig:xor_73} and \cref{fig:xor_80}, shows that through the spectral loss defined in \cref{eq:gain}, we found materials with more distinguishable $`0`/`1`$ responses. Although this result is promising, further investigation is needed to make a valid conclusion.

\section{Gradient-free Optimization Results}
\label{sec:EAresults}
\begin{figure}[ht]
\vskip 0.1in
\begin{center}
\centerline{\includegraphics[width=\linewidth]{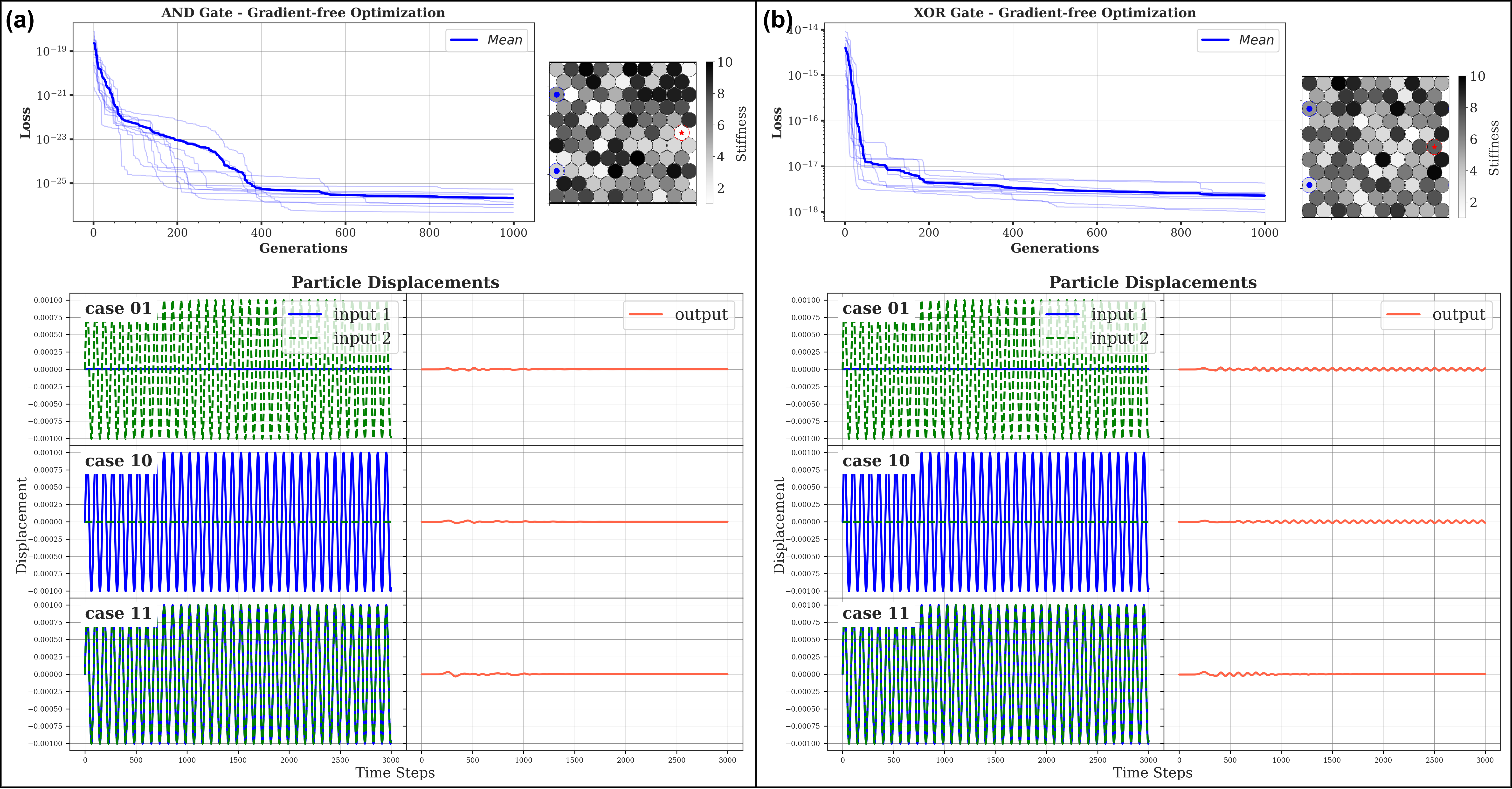}}
\caption{Gradient-free optimization of acoustic logic gates with the objective function defined in the time domain (\cref{eq:MAEloss}). The experimental setup and material properties are the same as the gradient-based optimization in \cref{sec:logicgate}. (a): acoustic AND gate. The output resembles the AND functionality in the \textit{$01$} and \textit{$10$} cases but not in the \textit{$11$} case. (b): acoustic XOR gate. The output in the \textit{$11$} case matches the output of an XOR function but the amplitude of vibration is negligible in the other two cases.}
\label{fig:EA_t}
\end{center}
\vskip -0.1in
\end{figure}

\begin{figure}[t]
\vskip 0.1in
\begin{center}
\centerline{\includegraphics[width=\linewidth]{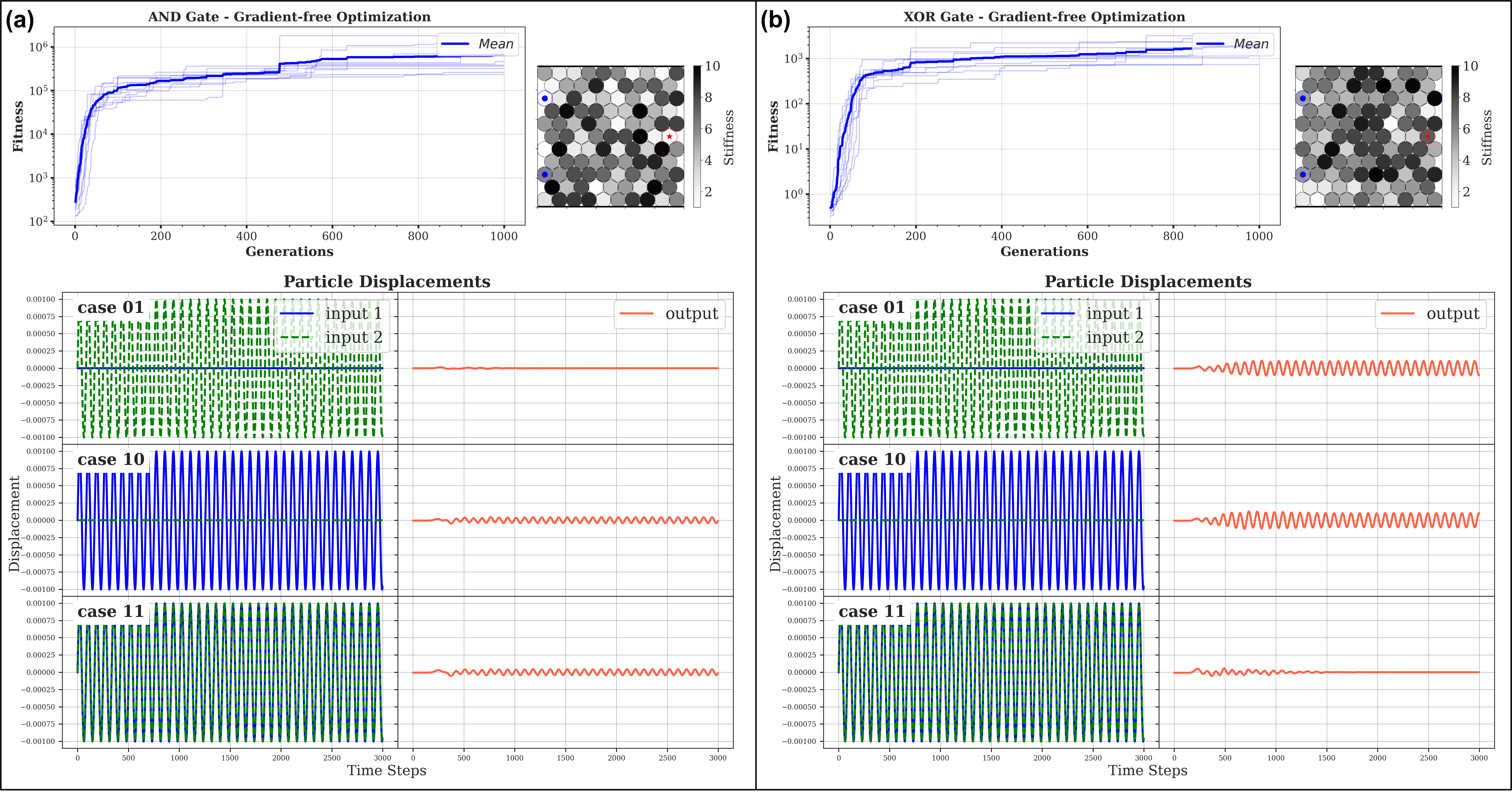}}
\caption{Gradient-free optimization of acoustic logic gates with the objective function defined in the frequency domain (\cref{eq:gain}). The experimental setup and material properties are the same as the gradient-based optimization in \cref{sec:logicgate}. (a): acoustic AND gate. (b): acoustic XOR gate.}
\label{fig:EA_f}
\end{center}
\vskip -0.1in
\end{figure}

In this section, we provide the results of the gradient-free optimization method discussed in \cref{sec:EA}. We considered two cases with the loss function defined in the time domain (\cref{eq:MAEloss}) and in the frequency domain (\cref{eq:gain}). \cref{fig:EA_t} and \cref{fig:EA_f} show the results of each experiment. these results are discussed in the main text (\cref{sec:discussion}).

\end{document}